# Is the use of Deep Learning and Artificial Intelligence an appropriate means to locate debris in the ocean without harming aquatic wildlife?

Zoe Moorton[1], Zeyneb Kurt [1*], Wai Lok Woo[1]

[1] Department of Computer and Information Sciences,

University of Northumbria, Newcastle Upon Tyne, UK

*Corresponding author: zeyneb.kurt@northumbria.ac.uk

**Abstract**

With the global issue of plastic debris ever expanding, it is imperative that the technology industry steps in. This study aims to assess whether deep learning can successfully distinguish between marine life and man-made debris underwater. The aim is to find if we are safely able to clean up our oceans with Artificial Intelligence without disrupting the delicate balance of the aquatic ecosystems.

The research explores the use of Convolutional Neural Networks from the perspective of protecting the ecosystem, rather than primarily collecting rubbish. We did this by building a custom-built, deep learning model, with an original database including 1,644 underwater images and used a binary classification to sort synthesised material from aquatic life. We concluded that although it is possible to safely distinguish between debris and life, further exploration with a larger database and stronger CNN structure has the potential for much more promising results.

# Introduction

## Background

It's no secret that marine debris has a devastating effect on marine life as mammals, sea birds and larger fish are frequently found entangled within nets and other materials, often ingesting the litter that is mistaken as prey or even just an object to play with. (D'Aurelio, 2019), (McAdam,R. 2017), (Day, R. 1980), (Bjornadal, K. et al.1994).

This results in lethal consequences as plastic fragments absorbs toxic materials and are highly contaminated with biphenyl polychlorinated (PCB), heavy metals and other noxious substances, if this does not poison the animal, it will cause alternative repercussions including reproductive disorders, hormone changes, higher disease risk or most commonly, obstructing the gastrointestinal tract, resulting in starvation and fatality. (D'Aurelio, 2019), (Mato, Y. et al. 2001).

McAdam (2017) states that you can "think of any aquatic setting, either local or exotic, and you can be sure that plastic has been found there" – even working its way into salt marshes (Viehman, S. et al. 2011). With that in mind, it is a global dilemma as micro plastics are now being discovered within human consumption, conveying the extensive scope of the problem (Sharma, S. et al. 2017). For almost a century, we have not successfully targeted this issue.

The overwhelming research of man-made debris within our waterways is rapidly proving that it is the largest predicament of our generation, with the Great Pacific Garbage Patch reaching the size of Mongolia, and unfortunately the ocean's four other gyres (rotating currents) are forming more garbage patches (NOAA, 2021) (National Geographic, 2019), it is crucial we begin to use technology to undertake this challenge.

Marine animals become trapped within this debris and quickly disappear, either from being consumed by predators or simply from sinking; making detection and sampling an incredibly difficult and limited task. "As a result, most data on entangled animals at sea are opportunistic anecdotal records" (Laist, D. 1997), confirming that we do not even know how many animals die in this way.

In the last two decades alone, hundreds of cetaceans have been killed or severely injured by marine litter. (Agence France-Presse. 2011). As a total average, more than 100,000 turtles, birds, seals, whales and dolphins are killed every year by discarded fishing equipment.

Whales are key contributors to the ecosystem of phytoplankton, who "contribute at least 50 percent of all oxygen to our atmosphere" and capture "an estimated 40 percent of all [Carbon Monoxide] CO produced." This means that phytoplankton act as the equivalent of 1.7 trillion trees or four Amazon forests (Chami, R. et al. 2019).

Although this piece of research is a very small aspect of the bigger picture, it has the potential to be a vital part of the first solution to cleaning up our oceans and water systems, protecting our rapidly declining marine life and potentially even human health.

The aim is to successfully use deep learning methods to distinguish between synthetic debris and aquatic life by using the following objectives:

- Compare current research on using artificial intelligence to identify or track synthetic materials in our water systems
- Test the method with the highest accuracy and efficiency is able to distinguish between marine life and material waste

The application of Machine Learning to detect and classify litter, is still very minimal but far outweighs the efficiency of current methods which are hugely time consuming and extremely limited. They also require manual labour and are unfortunately susceptible to human error. Being one of the most important environmental concerns of our time, it is critical we invest more research into Deep Learning as a progressive solution.

**Gaps in the Current Literature**

Most studies on the application of Artificial Intelligence have neglected to consider the safety and precautions involved with clearing the ecosystem without causing further harm.

Interestingly, popular categories studied in the oceans are plastic bottles, bags and food packaging, however, disregarded fishing nets cause a high fatality and are responsible for the greatest percentage of plastic in the ocean (Seaspiracy, 2021). Furthermore, many studies only considered floating plastics, which is unmeasurable and contributes to a very minimal percentage of marine litter.

Although the problematic debris is found near the surface where creatures mistake it for food or something to play with, therefore ingesting it (D'Aurelio, 2019), (McAdam,R. 2017), most of the ocean's plastic is in micro form and found in the deepest parts. (Peng, X et al. 2018) Moreover, only 1% of marine litter floats (Condor Ferries, 2021).

Most algorithms in current research are only looking at three or four separate categories, not only are these categories (usually including plastic bags, bottles and straws) representing a tiny portion of synthetic debris (for example straws only account for 0.03%. Seaspiracy, 2021), but also this does not accurately reflect the amount of degradation and distortion from wear, being underwater and also due to sea conditions (light, density, turbulence).

There is also a vast quantity of conditions that the algorithm would have to work under, with visibility being the most important factor. The authors in some of the papers are in fact currently testing this problem in real life ocean scenarios; however, it will still need further development, particularly in rivers which have completely different conditions and are responsible for approximately 1.15-2.41 million tonnes of plastic entering our oceans. (C.M., L et al. 2017)

However, de Vris, R. et al. (2021) recent work on the Great Garbage Patch has proven to successfully use artificial intelligence to track parts of the patch by collecting photographic data with a GPS enabled GoPro from above the waterline. Using a CNN to detect debris, they were able to survey the litter and map out the garbage patch in better detail.

**Related Work**

The paper on using deep learning to detect floating debris by Kylili, K. et al. (2018) compared training four network architectures (including R-CNN, YOLOv2, Tiny-YOLO & SSD) and concluded that Faster R-CNN had the highest accuracy, at the compromise of speed. YOLOv2 had an accuracy that was close to the R-CNN results, but it performed faster. They discovered that marine debris can be detected in real time using deep learning visual object detection methods - although this was from the surface and the authors do express their belief that it could be used underwater if the data limitations are overcome.

de Vris, R. et al. (2021) recently studied whether they could trace macro debris location and transportation. They used a two-fold approach, testing both the Faster RCNN and YOLOv5 architectures. The authors found that YOLOv5 outperformed FRCNN with the quantity of objects detected and with the smallest object size, however, they report this could be due to the hyperparameter settings they applied. They go on to explain that YOLOv5 only needed minimal changes for better performance and that both network architectures could have been further improved with more optimization. They produced successful results with what they describe as "the first real-world demonstration of a large-scale automated camera transect survey of floating marine litter", however due to limitations (such as that they are currently unable to detect anything smaller than macro debris), they believe their work is a strong framework for a methodology, rather than an example of findings.

Another experiment aiming to track and identify floating marine debris compared the use of CNN with a Bag of Features method. (Sreelakshmi, K, et al. 2019) The CNN used Convolutional and Bottleneck Layers achieving 77.5% accuracy when classifying. However, the Bag of Features method only achieved a 62.5% accuracy (using MatLab & SURF features).

Comparing the two results clearly indicates that CNN was more accurate with an added benefit of speed - it took half the computational time that the Bag of Features needed.

Another method explored using Long Short Term Memory Network (LSTM) with Cross Correlations and Association rules (Apriori) to identify characteristics of water pollutants, they discovered that the water quality correlation maps they produced were in fact able to accurately identify fluctuations effectively. Whilst this is useful to detect where large quantities of litter are within the ocean, this would work better as a technology used in addition to a Deep Learning algorithm, to encourage accurately pinpointing our largest problematic areas. (Wang, P. 2019). Particularly as they have claimed that their Artificial Intelligence scheme could potentially work within aquatic systems.

The paper (Chazhoor et al, 2021) benchmarked the three widely implemented architectures on the WaDaBa dataset to find out the best model with the support of transfer learning. To ease the recycling process worldwide, seven different types of plastic have been categorized based on their chemical composition. These are Polyethylene Terephthalate (PET or PETE), High-Density Polyethylene (HDPE), Polyvinyl Chloride (PVC or Vinyl), Low-Density Polyethylene (LDPE), Polypropylene (PP), and Polystyrene (PS or Styrofoam). PET, HDPE PP and PS dominate the household waste and segregating them into their respective types will allow the reuse of certain types and recycling of other types of plastics. The paper is the first benchmark paper aimed towards classifying different types of plastics from the images using deep learning models, and this has stimulated the research in this area and serve as a baseline for future research work.

To conclude, the current algorithms available are successful and effective; however, comparing the studies, the results clearly indicate that CNN was the most accurate and fastest approach. Convolutional Neural Networks (CNN) can train themselves on debris and continue to learn using transfer learning, concluding in highly accurate and efficient results. (Kylili, K. et al. 2018)

Within this research, we have collated an extensive and original database of not only litter but also aquatic life with an emphasis on submerged imagery that covers a diverse and comprehensive range of data. Incorporating different sea conditions, light levels and depths, as well as all varieties of synthesised materials and in various stages of degradation (or other distortions). This is important to ensure that the CNN does not just learn what is correct but also what is incorrect.

The proposed architecture is especially unique in its binary classification - we split all the data into two simple categories of what is 'living' and what is not - putting the protection of the ecosystem and of life as the priority and trained the system how to differentiate between the two with a current accuracy of 89%. Therefore, picking up most synthesised matter and avoiding (protecting) aquatic life.

It also covers a broader spectrum of items rather than focusing on specific categories at a specific depth, Therefore, training it to recognize properties of either debris or life, rather than single items. We have used data augmentation to diversify the different aspects of images as well as increasing the number of images to deal with class imbalance issues.

## Materials and Methods

### Database Collection

A diverse range of organizations and dive centres collaborated with us by providing their images.

The main database acquired was retained from Japan Agency for Marine-Earth Science and Technology (JAMSTEC). JAMSTEC has a huge database of submarine images and footage of deep-sea debris in Japan. This was perfect for different depths, density and light but also for the variety of objects within the photos, with different levels of degradation.

### Data Augmentation Process

On our dataset, only a small amount of data augmentation was used by manually cropping and rotating any images that were particularly rich in information; mostly in instances where both categories (such as a fish and a plastic bottle) were in one image; that image would then be split into two separate images, individually cropping the items. We started with 1,318 photos, after augmentation was applied, our training set consisted of 1,644 images and 100 images for the testing set.

**Sorting Data**

Firstly, we ensured a high-quality collection. In some cases, the image quality was too poor, so it was decided it would benefit the system by removing them.

Similarly, to poor quality, anything irrelevant was also removed, so as not to confuse the algorithm. In the case of our dataset, it was mostly the JAMSTEC submarine machinery parts that were in the way, which was resolved by manually cropping the images.

The biggest issue that arose was the strange (yet incredible) phenomenon of how much aquatic life had chosen to inhabit pollutants. An ethical dilemma therefore arises, as to not disturb the sensitive and delicate balance of the ecosystem, by removing these items (Fulton, M. et al. 2019). We chose to remove any such images to prevent confusion when training the network architecture.

From sorting the dataset, it was often questioned whether the Convolutional Neural Network would be able to safely distinguish between very commonly confused items. For example, a jellyfish and a plastic bag closely resemble each other. There were also cases of starfish, that as researchers, we ourselves were not sure whether they were real or plastic. Therefore, there is a very real danger that a neural network could make the same mistake.

**Training a Model for a Binary Classification Task**

In this study, classes are defined as follows: an image which has man-made debris in it ('Litter' or '1'), or it has natural aquatic life in it ('Animals' or '0'), this class can include plants or any variety of marine biology.

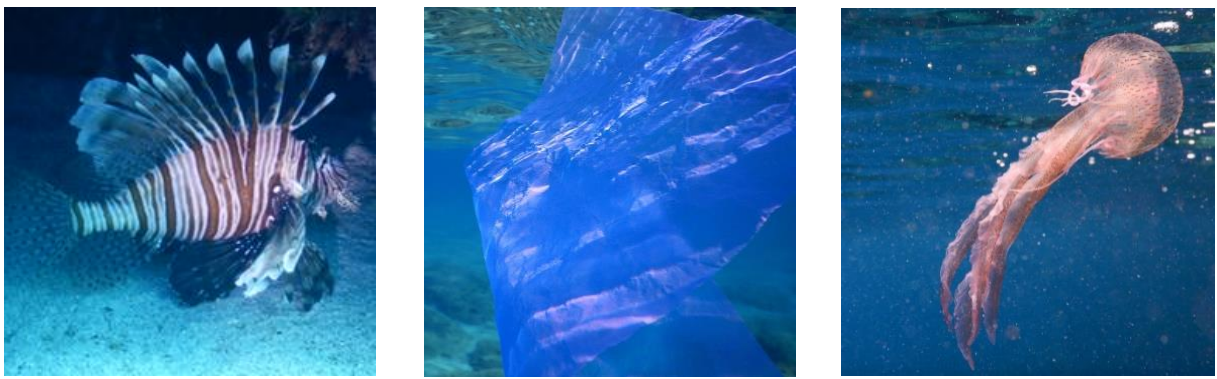

Figure 1: Examples of database images. Photo Credit: Maria Shokouros-Oskarsson

For now, we chose not to use subclasses, in order to test this version first and see if it is successful. In further tests, a training set of multiple categories would be considerably valuable.

**Network Architecture**

We chose Convolutional Neural Networks as they can handle a large quantity of data and have a high-performance rate in the ever-developing machine learning industry. CNN algorithms are the strongest neural network for image recognition. The system learns to strategically, on different layers, detect the lines and edges in the input image. It is then able to distinguish between them, identifying segments of an image resulting in feature extraction.

A major disadvantage to deep learning algorithms having many parameters, and the use of Max Pooling, is the higher risk of overfitting (Guo, Y. et al. 2016). As this is an entirely new architecture with a brand new, untrained dataset, it required a large amount of trial and error of adjusting the different parameters to avoid overfitting. To closely monitor the progress of the architecture, Tensorboard was used, so that we could frequently compare graphs on where overfitting was occurring and make appropriate adjustments.

All images were resized, to a width and height of 140px, using Python to maintain a uniform size throughout all images and ensure that parts would not be cropped out. Additionally, 140px when tested, showed enough detail to recognize the shapes and patterns within the image but was not too high to slow down the run time.

In the case of using such a vast variety of underwater debris and aquatic life, it was felt that the most information necessary should be used, even if it compromises speed; so three channel values RGB were used, instead of one.

To eliminate the issue of class imbalance, the dataset has been split to have exactly 50% of each category. Therefore, out of the 1,644 training images, 822 were in each category. An additional precaution taken, was to randomly shuffle the data, as the system would otherwise learn that the data is in order and again, guess appropriately.

This system was created on a Dell Inspiron i7-7700HQ CPU 2.8GHz, 16GB RAM, 64-Bit with a Nvidia GeForce GTX graphics card, using Windows 10.

The algorithm was programmed using Python in PyCharm and the specs of the system used were: Tensorflow, Keras and ReLU. Using cv2, we imported the collected and built dataset, created 2 x 2 Windows, extracted the Max Pooling value and used the Adam optimization.

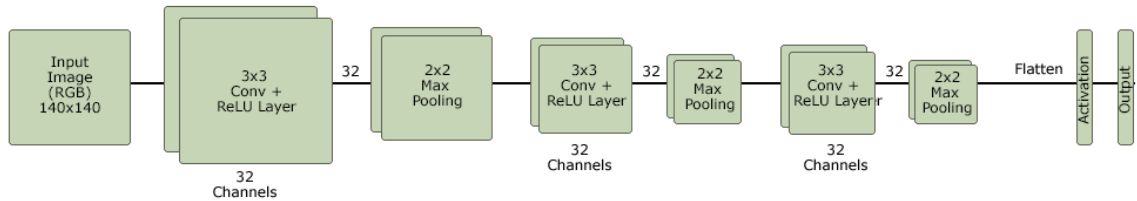

Figure 2: Block Diagram of Proposed Architecture

This resulted in the architecture having the following parameters: Three Convolutional Layers with 32 Nodes and No Dense Layers, the second Kernel size was reduced to a (2, 2) window, the validation split had to be reduced to 0.1 and it ran 95 epochs. As shown in Figure 2. We used the Rectified Linear activation Unit (ReLU) as the activation function and the last layer used the logistic sigmoid function to enable binary classification. The validation split is usually optimized at 20% (0.2) of the set, however, this architecture would favour one of the classifications, if at this setting.

When the final architecture was confirmed, we ran it 10 times to take an average of the Validation Loss, Validation Accuracy, Training Loss and Training Accuracy shown in Table 1, intermittently between rerunning the architecture, we continued to run Tests against Training Images and Test Images, to ensure it was classifying correctly still.

| Train Loss | Validation Loss | Train Accuracy | Validation Accuracy |
|---|---|---|---|
| 0.23 | 0.56 | 0.90 | 0.72 |
| 0.21 | 0.55 | 0.90 | 0.73 |
| 0.19 | 0.67 | 0.92 | 0.73 |
| 0.22 | 0.52 | 0.90 | 0.75 |
| 0.16 | 0.55 | 0.93 | 0.77 |
| 0.18 | 0.84 | 0.92 | 0.67 |
| 0.20 | 0.73 | 0.92 | 0.74 |
| 0.19 | 0.57 | 0.93 | 0.75 |
| 0.21 | 0.64 | 0.90 | 0.73 |
| 0.21 | 0.63 | 0.91 | 0.74 |
| 0.20 | 0.63 | 0.91 | 0.73 |

Table 1: Training Results

**Results, Analysis and Evaluation**

The original aim of this research was to test if we could successfully train a CNN framework to distinguish between underwater debris and aquatic life, safely and to an accuracy of at least 85%, as we collected a new database.

|  n = 100 | PREDICTED: Animal | PREDICTED: Litter | |
|---|---|---|---|
| ACTUAL: Animal | TP = 47 | FN = 8 | 55 |
| ACTUAL: Litter | FP = 3 | TN = 42 | 45 |
|  | 50 | 50 | |

Figure 3: Confusion Matrix

The test results have shown that from a set of 100 tested images, 89% accurately classified whether there was marine life or debris.

The obtained result shows promising signs of being able to adapt to a much larger scale database, with more detail and a stronger CNN architecture.

There were limitations to this study, as finding publicly accessible underwater images of debris proved particularly difficult, and the images that were donated to us were not already labelled, so we had to complete this manually, which was time consuming.

We also found similar challenges with de Vris, R. et al. (2021), that many images were not of a high enough quality, consequently reducing our database to a size that was not suitable for CNN training. However, after the use of data augmentation, we were successfully able to create a new database that had a high accuracy level. From here we believe this research is therefore, a strong preliminary methodology for future research in this field with a larger database and a more powerful solution that can allow further depth and contain more classification groups.

This research will be publicly accessible, as it is unique in its type that it is one of the first papers that not only is looking for a feasible solution for our polluted water crisis but also has an emphasis on the perspective of animal welfare and ecological conservation. Therefore, it has also collated a large quantity of thoroughly thought-out scenarios and information that future papers will hopefully prioritise their studies and development with.

**CONCLUSION AND RECOMMENDATIONS**

This research sets out to answer the question: can deep learning successfully distinguish between marine life and man-made debris underwater? From the research we conducted, the authors believe that yes, we can plausibly use image recognition to safely remove debris underwater without disrupting aquatic ecosystems. Kylili, K. et al. (2018) quite rightly expressed concern about using image recognition underwater with the vast variety of debris available, however we have proven that subclasses are possible to use as a base to start with, and with future development on this research, we believe this could develop into a working large scale research.

While this research scratches the surface of potentially using Artificial Intelligence to distinguish successfully and safely between life and man-made debris; there are plenty of other factors that should be considered and researched moving forward.

Although this research was diverse in its nature of depth, geographic location, and sea conditions, it did not cover a wide enough variety and would need to include a database that includes these variables or considers fresh waterways and the aquatic life that lives there.

Additionally, this structure was only trained on matter that exceeded the size of micro plastics (microplastics are <5 mm). Therefore, it remains untested on any debris or living creatures under that size.

Future research should be tested for image segmentation to ensure that biological entities (such as plants or animals) attached onto debris, are not retrieved.

Other variables would be to diversify the variety of species of all underwater creatures, such as how would this machine specifically respond to stingrays embedded within the sand or coral with plastic wrapped on it.

Additionally, an entire classification could be dedicated to entangled sea creatures and plants, potentially then essentially reporting back to the user of an endangered life.

Further considerations include asking if it would be able to handle more complex situations? Such as accurately pinpointing floating plastic in a sea of jellyfish. Although de Vris, R. et al. (2021) at Ocean Cleanup are doing a fantastic job of mapping patches of debris within the gyre, could we test our method on retrieving large quantities of underwater debris without picking up small life within something as vast and compact as the Great Garbage Reef?

It would also be valuable for the algorithm to be able to continue unsupervised learning with positive and negative reinforcement, so that it constantly trains itself and continues to improve its accuracy.